%% file: imu_mono_mfi2017_arxiv.tex
\documentclass[letterpaper, 10 pt, conference]{ieeeconf}  

\IEEEoverridecommandlockouts                              

\usepackage{amsmath,amssymb,mathtools}
\usepackage{scalerel,stackengine}
\usepackage{upgreek}
\usepackage{bm}
\usepackage{graphicx}
\usepackage{epsfig} 
\usepackage{cite}
\usepackage[shadow,textsize=footnotesize,color=yellow,colorinlistoftodos]{todonotes}

\usepackage{dcolumn,booktabs}
\newcolumntype{d}[1]{D{.}{.}{#1}}
\newcommand\mc[1]{\multicolumn{1}{c|}{#1}} 

\newcommand\equalhat{\mathrel{\stackon[1.5pt]{=}{\stretchto{%
  		\scalerel*[\widthof{=}]{\wedge}{\rule{1ex}{3ex}}}{0.5ex}}}}

\newcommand\Tstrut{\rule{0pt}{2.2ex}}         

\graphicspath{{figs/}}

\title{\LARGE \bf
A Loosely-Coupled Approach for Metric Scale Estimation in Monocular Vision-Inertial Systems
}

\author{Ariane Spaenlehauer\thanks{Sorbonne Universit{\'{e}}s, universit{\'{e}} de technologie de Compi{\`{e}}gne, CNRS, UMR 7253 Heudiasyc-CS 60 319, 60 203 Compi{\`{e}}gne Cedex, France (e-mail: {\scriptsize \textsf{ariane.spaenlehauer@hds.utc.fr}}, {\scriptsize \textsf{vincent.fremont@hds.utc.fr}}, {\scriptsize \textsf{asekerci@ieee.org}}, {\scriptsize \textsf{isabelle.fantoni@hds.utc.fr}}).} $\quad$$\quad$  Vincent Fr\'{e}mont $\quad$$\quad$  Y. Ahmet \c{S}ekercio\u{g}lu $\quad$$\quad$  Isabelle Fantoni}

\begin{document}

\maketitle
\thispagestyle{empty}
\pagestyle{empty}

\begin{abstract}

  In monocular vision systems, lack of knowledge about metric distances caused
  by the inherent scale ambiguity can be a strong limitation for some
  applications. We offer a method for fusing inertial measurements
  with monocular odometry or tracking to estimate metric distances in
  inertial-monocular systems and to increase the rate of pose
  estimates. As we performed the fusion in a loosely-coupled manner,
  each input block can be easily replaced with one's preference, which
  makes our method quite flexible. We experimented our method using
  the ORB-SLAM algorithm for the monocular tracking input and Euler
  forward integration to process the inertial measurements. We chose
  sets of data recorded on UAVs to design a suitable system for
  flying robots.

\end{abstract}
\section{INTRODUCTION}

In recent times, research interest for monocular vision has been
strongly increasing in robotics applications. The use of vision-based
sensors such as cameras have numerous advantages. They have low energy
consumption, they can be manufactured in very small sizes and their
cost is dramatically reducing every year. Their typical applications
include autonomous navigation, surveillance or mapping. A key issue
that directly impacts on the success of these applications is the
estimation of locations and distances by using the information
gathered by these visual sensors. Several studies show that
combining visual information with low-cost, widely available
inertial sensors, Inertial Measurement Units (IMUs), improves
the accuracy of these estimations.

In this paper, we focus on this kind of sensor sets, called
``inertial-monocular'' systems, which are composed of a monocular
camera and an IMU attached to Unmanned Aerial Vehicles (UAVs). We
present a computationally lightweight, and fast solution for
estimating the metric distances over the visual information collected
by the monocular camera of a UAV. The problem is summarized as
follows: By using the frames provided by the camera, algorithms for
odometry or Simultaneous Localization And Mapping (SLAM)
\cite{cadena2016past} can estimate the camera positions and
orientations (camera poses) and, for the SLAM, create a 3-D
representation of the environment. However, the estimates are
calculated up to scale \cite{hartley2003multiple}. This scale
ambiguity is inherent to monocular vision and cannot be avoided. When
a 3-D scene is captured by the camera and projected into a 2-D frame,
depth information is lost. By measuring the same scene from different
points of views, depth can be reconstructed up to scale. The scale
factor is different for each frame, nevertheless recent algorithms
provide consistent camera pose estimates, which include the estimation
of this scale factor. However, the estimation of the scale factor does
not provide metric distances. The scale factor is used to ensure
consistency in the estimation of camera positions, i.e., large
distances in the world coordinate frame measured from a frame $F_{i}$
remain large even if they are measured again from another frame
$F_{j}$. To recover metric distances, the length of the camera
position vector has to be rescaled using a coefficient. This scaling
operation results in metric estimates for distances. The aim of our
method is to compute this scaling coefficient.

As mentioned above, monocular vision systems cannot recover the scale
of the world; therefore, at least one additional sensor capable of
measuring or estimating metric distances must be added to the
system. Several sensors can meet this requirement such as lidar,
ultrasound or IMU. The use of IMU is often
preferred in UAVs because of its small size and low cost. However, IMU
does not measure distances directly but acceleration and angular
velocities in the inertial frame. Distances can be recovered through
the calculation of positions by integrating the acceleration
measurements but, consequently, the estimates drift quickly with the
error accumulation, which prevents any long-term integration.

\begin{figure}[h!]
  \centering
  \includegraphics[width=0.9\columnwidth]{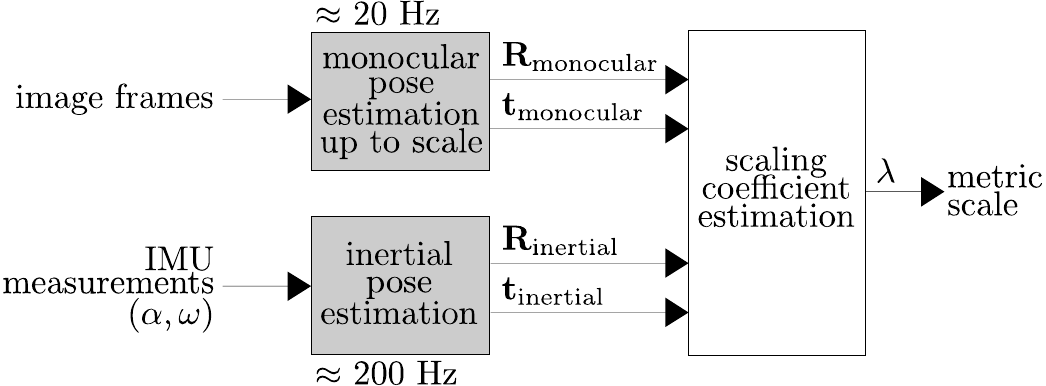}
  \caption{Overview of the system architecture: The blocks in grey can
    be replaced with one's preferences. For our experiments, we used
    the ORB-SLAM algorithm for monocular pose estimation and Euler
    forward integration for inertial pose estimation. }
  \label{fig:metric_scale}
\end{figure}

The approach we propose is based on distances ($L^{2}$ norm of
translation vectors) and is suitable to fuse the output of any
monocular odometry or the tracking part of SLAM algorithms with
inertial measurements. An overview of the system architecture is shown
in Fig. \ref{fig:metric_scale}. In the following sections, we first
provide an overview of the leading approaches. Then, mathematical
details of the estimation of scaling coefficient by using IMU
measurements are presented. Finally, we test the validity of our
method over a set of UAV trajectories \cite{Burri25012016}.

\section{RELATED WORKS}
Two main approaches for monocular visual-inertial fusion can be
distinguished in the literature: Loosely-coupled filtering
\cite{weiss2011real} \cite{hu2014sliding} \cite{sa2013monocular} and
tightly-coupled systems \cite{mur2017visual} \cite{concha2016visual}.


In tightly-coupled
approaches, the fusion is done at a low level of the
system. Therefore, this requires a deep understanding of the involved
algorithms and specific design for the system.

The method described in \cite{concha2016visual} is the inertial
extension of the DPPTAM \cite{concha2015dpptam}, a direct SLAM
algorithm. The tracking thread is modified to include the IMU
measurements. The Gauss-Newton optimization is used to minimize the
intensity and IMU residuals. The state vector is composed of the
position, orientation and velocity of the robot and the IMU
biases. The IMU measurements are integrated between two consecutive
keyframes. The IMU residuals are the error of the inertial integration
between two keyframe with regard to the state value at the
corresponding time. The intensity residuals are the photometric error
between two keyframes. They are calculated by reprojecting the map
points in the keyframes using the estimate of the relative camera
pose. The optimization of both residuals provide the final pose
estimate of the current keyframe with regard to the world coordinate
frame.

The method described in \cite{mur2017visual} is the inertial extension
of ORB-SLAM \cite{mur2015orb}. In ORB-SLAM, no functionality is
provided to calculate the uncertainty of pose estimates. Therefore,
the implemented method needs to avoid the direct use of the
uncertainty of the camera pose. To represent information about
uncertainty, the authors use information matrices computed either from
the preintegration of the IMU measurements or from the feature
extraction. The reprojection error and inertial error are minimized
using the Gauss-Newton optimization. The reprojection error comes from
the reprojection of map points in the current keyframe while the IMU
error is derived from the preintegration equations described in
\cite{forster2015imu}.

In contrast to tightly-coupled approaches, in loosely-coupled approaches, the vision part is considered as a
black box, only the output of the box is used. In most loosely-coupled
algorithms such as \cite{weiss2011real} \cite{hu2014sliding}
\cite{sa2013monocular}, the filter, which fuses the measurements, is
derived from Kalman Filtering, e.g., Extended Kalman Filter or
Multi-State Constraints Kalman Filter. The state is, at least,
composed of the position, orientation, velocity and biases of the
IMU. The differential equations, which govern the system and the IMU
measurements, are used to predict the state. The incorporation of
monocular visual measurements is done through the measurement model
when the Kalman gain needs to be computed (the visual measurements
update the state when the innovation is calculated).

In \cite{weiss2011real}, the state vector additionally includes the
calibration states (the constant relative position and orientation
between the IMU coordinate frame and the camera coordinate frame) and
a failure detection system. When a failure is detected (abrupt changes
in the orientation estimates with regard to the measurement rate), the
related visual measurements are automatically discarded to prevent the
corruption of data.

In \cite{hu2014sliding}, the authors use trifocal tensor geometry
which considers epipolar constraints in triples of consecutive images
instead of pairs of images. Therefore, in addition to the usual IMU
states, the state vector also contains the pose and orientation of the
two previous keyframes.

In \cite{sa2013monocular}, the fusion is done by using measurements
from three sensors: In addition to the visual and inertial sensors, a
sonar is included in the system to measure distances (the altitude
between the UAV and the ground).  IMU is used to detect whether the
UAV is flying level or tilted. If it is level, the sonar measurements
are directly used to estimate the altitude. Otherwise, IMU
measurements help to rectify the incorrect altitude information due to
the tilting of the UAV. The scale factor estimation is represented as
an optimization problem between the sonar and visual altitude
measurements which is solved using the Levenberg-Marquardt algorithm
\cite{more1978levenberg}.


In this paper, we propose an approach for fusing monocular and
inertial measurements in a loosely-coupled manner which is simple to
implement, requires small computational resources and so, is suitable
for UAVs. We decided to design a loosely-coupled approach to make the
methods used for visual tracking and IMU measurement integration easy
to replace with any other method ones may prefer, which ensures better
flexibility and usability for our approach. In our studies, we used
the ORB-SLAM algorithm \cite{mur2015orb} for the visual tracking part
and Euler forward integration for the inertial measurements
processing.

\section{SCALING COEFFICIENT ESTIMATION WITH IMU MEASUREMENTS}

\subsection{Coordinate Frames}
Our system (see Fig. \ref{fig:metric_scale}) is composed of two sensors (a camera and an IMU) attached on a rigid flying body, the UAV. We distinguish four coordinate frames:  camera \{$C$\}, vision \{$V$\}, inertial \{$I$\} and world \{$W$\} coordinate frames\footnote{The symbols used in the following paragraphs are given in
Table \ref{tab:symbols}.}. The IMU measures data in \{$I$\} attached to the body of the UAV. The integration of IMU measurements results in the estimation of the pose of the IMU in \{$W$\}. The monocular pose estimation algorithm outputs the camera poses in \{$V$\}, which corresponds to the first \{$C$\} coordinate frame when the tracking starts, i.e.
\begin{equation}
\{{V}\} \equalhat \{{C}\}_{\text{t}=0}
\end{equation}
The matrix $^{I}\mathbf{T}_{C}$, which represents the transformation
between \{$C$\} and \{$I$\}, is constant, and can be computed off-line
using a calibration method \cite{Mirzaei2008}. In the EuRoC dataset
sequences that we used for our experiments, $^{I}\mathbf{T}_{C}$ is
already provided. We consider that \{$W$\} corresponds to \{$I$\} at
the moment tracking starts, when the \{$V$\} coordinate frame is
generated, so
 \begin{equation}
 \{{W}\} \equalhat \{{I}\}_{\text{t}=0}
 \end{equation}
 In the following paragraphs, we consider that the monocular pose estimation algorithm outputs measurements in the world coordinate frame by applying the formula
\begin{equation}
{}^{W}\!\mathbf{p} = {}^{I}\mathbf{T}_{C} {}^{V}\!\mathbf{p}
\end{equation}
where $\mathbf{p}$ is a 3-D measurement by the monocular pose estimation algorithm.

\subsection{Integration of IMU Measurements}

\begin{table}
  \caption{Mathematical Notation}
  \centering\input{symbols-notation}
  \label{tab:symbols}
\end{table}

\begin{figure}[h!]
  \centering\includegraphics[width=\columnwidth]{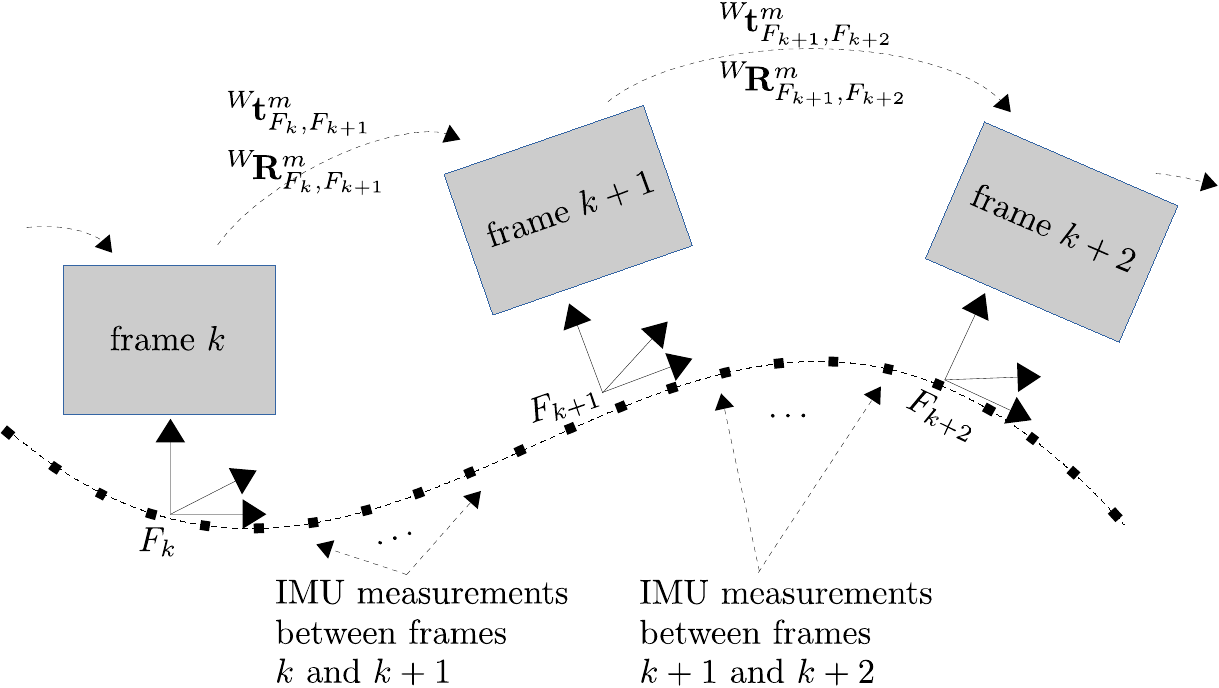}
  \caption{IMU measurements obtained between the two subsequent image
    frames are used to calculate the translation vectors and rotation
    matrices.}
  \label{fig:imu_measurements}
\end{figure}

We assume having a monocular pose estimation algorithm which outputs
consistent camera poses in the world coordinate frame. As a monocular
camera cannot be used to calculate metric distances, we use the IMU
measurements to compute the scaling coefficient. IMU is a sensor
composed of three accelerometers and three gyroscopes which
respectively measure accelerations and angular velocities along each
of the three axis of the inertial frame. The pose of the IMU can be
estimated by integrating the accelerometer and gyroscope
measurements. However, the integration of the IMU measurement noise
and the IMU biases makes the pose estimates to drift fast. Therefore,
IMU measurements must be integrated only over a short period for
limiting the drift and consequently to corrupt the estimates. We
integrate IMU measurements between two consecutive image frames. So,
if $F_{k}$ and $F_{k+1}$ are two coordinate frames associated with
image frames $k$ and $k+1$ (see Fig. \ref{fig:imu_measurements}), the
integration results in the estimation of the rotation matrix
$^{F_{k}}\textbf{R}_{F_{k+1}}$, velocity
$^{W}\textbf{v}_{F_{k}, F_{k+1}}$ and translation
$^{W}\textbf{t}_{F_{k}, F_{k+1}}$ vectors between the two consecutive
frames $F_{k}$ and $F_{k+1}$ in the world coordinate frame using IMU
measurements.

We integrated the IMU measurements using Euler forward integration following the description given in \cite{concha2016visual}
\begin{equation}
^{F_{i}}\textbf{R}_{F_{j}} = \prod_{p=k}^{k+N-1} \exp_{\text{SO(3)}}([\bm{\upomega}(p)+\mathbf{b}_{\omega}(p)]^{\wedge} \Delta T) 
\end{equation}
where $\bm{\upomega}$ is the vector of gyroscope measurements, $\mathbf{b}_{\omega}$ the gyroscope bias, $k$ the time step of frame $F_{i}$, $k+N$ the time step of frame $F_{j}$, $N-1$ the number of IMU measurements between the consecutive frames $F_{i}$ and $F_{j}$ and $\Delta T$ is the time step size (in seconds)
\begin{equation}
^{W}\textbf{v}_{F_{i}, F_{j}} = \sum_{p=k}^{k+N-1}[^{W}\mathbf{R}_{I_{p}}(\mathbf{a}(p)+\textbf{b}_{a}(p)) - \mathbf{g}] \Delta T
\end{equation}
where $^{W}\mathbf{R}_{I_{p}}$ is the rotation matrix between the world coordinate frame and the IMU coordinate frame at time $p$, $\mathbf{a}$ is the vector of accelerometer measurements, $\mathbf{b}_{a}$ the accelerometer bias and $g$ the gravity vector
\begin{equation}
  \begin{aligned}[b]
  ^{W}{\textbf{t}}_{F_{i}, F_{j}}  = & N ^{W}\mathbf{v}_{F_{i}}\Delta T +  \\
  & \frac{1}{2}\sum_{p=k}^{k+N-1}\!\! \Big[ \left( 2 \left( k+N-1-p\right)+1 \right) \\
  & \left({w}^\mathbf{R}_{p}\left(\mathbf{a}(p)+\mathbf{b}_{a}(p)\right) - \mathbf{g}\right)\Big] \Delta T^{2}
  \end{aligned}
\end{equation}
The $\exp_{SO(3)}$ operator maps an vector of $\text{so(3)}$ to a matrix of $\text{SO(3)}$. The wedge operator $.\wedge$ convert a $3\times1$ vector into an element of $\text{so(3)}$, i.e., a skew-symmetric matrix of size $3\times3$. The IMU biases, $\mathbf{b}_{a}$ and $\mathbf{b}_{\omega}$, were modeled as a random walk process
\begin{equation}
\mathbf{b}_{a}(k+1) = \mathbf{b}_{a}(k) + \Delta T \bm\upsigma_{a}^{2}
\end{equation}
where $\bm\upsigma_{a}^{2}$ is the variance associated to the IMU accelerometers
\begin{equation}
\mathbf{b}_{\omega}(k+1) = \mathbf{b}_{\omega}(k) + \Delta T \bm\upsigma_{\omega}^{2}
\end{equation}
where $\bm\upsigma_{\omega}^{2}$ is the variance associated to the IMU accelerometers

Note that the IMU integration equations (Eqs. 4, 5, 6) can be replaced with another approach for numerical integration (such as \cite{forster2015imu}) as long as this calculates the translation vector between two consecutive frames in the world coordinate frame \{$W$\}. 

It is assumed that the estimates provided by the monocular pose estimation algorithm $\mathbf{x}^{\text{m}}$ drift slower than the estimates $\mathbf{x}^{\text{i}}$ computed using the IMU measurements, i.e., $\mathbf{x}^{\text{m}}$ is more accurate than $\mathbf{x}^{\text{i}}$. At each incoming frame, we used the value of $\mathbf{x}^{\text{m}}$ to initialize $\mathbf{x}^{\text{i}}$. We observed that a good initialization for the IMU estimates can greatly improve the accuracy of the estimation of the scaling coefficient. However, the initialization of the estimates $\mathbf{x}^{\text{i}}$ is not discussed in this paper and is part of the further improvement we plan to do. 

We also benefit from the high measurement rate of the IMU (between 100 Hz and 200 Hz) to provide fast pose estimates. The pose estimates from the vision algorithm $\mathbf{x}_{\text{m}}$ can be updated once per new frame at maximum. Therefore, the camera pose is updated at the frame rate, usually around 30 Hz, which can be too slow for some applications such as control or navigation.

\subsection{Calculation of Scaling Coefficient}
We want to find the scaling coefficient $\lambda$ as follows
\begin{equation}
  \Vert ^{W\!}\mathbf{t}^{\text{i}} \Vert_{2} = \lambda \Vert ^{W\!}\mathbf{t}^{\text{m}} \Vert_{2}
\end{equation}
where $^{W\!}\mathbf{t}^{\text{i}}$ are the translation vectors of the camera position given by the integration of IMU measurements in the world coordinate frame, $^{W\!}\mathbf{t}^{\text{m}}$ are the translation vectors of the camera position given by the monocular odometry or SLAM algorithm in the world coordinate frame and $\Vert . \Vert_{2}$ is the $L^{2}$ norm of a vector.

For each incoming new frame $F_{j}$, the translation of the camera between the consecutive frames $F_{i}$ and $F_{j}$ in the world coordinate frame given by the monocular vision algorithm $^{W\!}\mathbf{t}^{\text{m}}_{F_{i}, F_{j}}$ is measured. We then integrate the corresponding IMU measurements using the Eq. (1), (2) and (3) to obtain the corresponding translation from the inertial measurements $^{W\!}\mathbf{t}^{\text{i}}_{F_{i}, F_{j}}$
\begin{equation}
\lambda_{F_{i}, F_{j}} = \frac{ \Vert ^{W\!}\mathbf{t}^{\text{i}}_{F_{i}, F_{j}} \Vert_{2}}{\Vert ^{W\!}\mathbf{t}^{\text{m}}_{F_{i}, F_{j}} \Vert_{2}}
\label{eq:lambda}
\end{equation} 
So Eq. \ref{eq:lambda} provides an estimated value of the scaling coefficient $\lambda$.

We can measure $\lambda$ for each frame, but the measurement noise on each measurement is significant because both the outputs of the SLAM algorithm and the IMU integration drift. Four methods have been tested to calculate the scaling coefficient $\hat{\lambda}$ using the measurements $\lambda_{F_{i}, F_{j}}$. The four methods are: moving average on $\lambda_{F_{i}, F_{j}}$ with an additive model for the error, moving average on $\log (\lambda_{F_{i}, F_{j}})$ with a multiplicative model for the error, an autoregressive Filter and a Kalman Filter.

The moving averages are calculated over the available measurements at time $t$
\begin{equation}
\hat{\lambda}_{1} = \frac{1}{M} \sum_{k=2}^{M-1} \left( \frac{\Vert ^{W\!}\mathbf{t}^{\text{i}}_{F_{k}, F_{k+1}}\Vert_{2}}{\Vert ^{W\!}\mathbf{t}^{\text{m}}_{F_{k}, F_{k+1}}\Vert_{2}} \right)
\end{equation}
where the error model is additive
\begin{equation}
\hat{\lambda}_{2} = \exp \left( \frac{1}{M} \sum_{k=2}^{M-1} \log \left( \frac{\Vert ^{W\!}\mathbf{t}^{\text{i}}_{F_{k}, F_{k+1}}\Vert_{2}}{\Vert ^{W\!}\mathbf{t}^{\text{m}}_{F_{k}, F_{k+1}}\Vert_{2}} \right) \right)
\end{equation}
where the error model is multiplicative and $M$ is the number of frames at the considered discrete time $t$. The first frame is skipped because the error on the first IMU measurement is generally large. 

We also decided to implement an autoregressive filter (AR) to estimate $\hat{\lambda}$. Moreover, this filter can be used to check whether the measurements are correlated in time. The current value of the filter output $y(i)$ is a weighted linear combination of the $p$ previous outputs ($p$ is the order of the filter) and the current measurement. The weights are computed by solving the Yule-Walker equations

\begin{equation}
y(i) = K + s(i) + \sum_{j=1}^{p}\alpha_{i} y(i-j)
\end{equation}
where $y(i)$ is the output of the AR filter at discrete time $i$, $\alpha_{i}$ are the weights calculated with the Yule-Walker equations, $s$ is a zero-mean random variable with
\begin{equation}
s(i) = \lambda(i) - K
\end{equation}
\begin{equation}
\lambda(i) = \frac{\Vert ^{W\!}\mathbf{t}^{\text{i}}_{F_{i}, F_{i+1}}\Vert_{2}}{\Vert ^{W\!}\mathbf{t}^{\text{m}}_{F_{i}, F_{i+1}}\Vert_{2}}
\end{equation}
The weights $\alpha_{i}$ and bias term $K$ are calculated solving
\begin{equation}
\begin{bmatrix*}
\mu \\ c_{1} \\ c_{2} \\ \vdots \\ c_{p}
\end{bmatrix*} = \begin{bmatrix*} 1 & \mu & \mu & \ldots & \mu \\ \mu & c_{0} & c_{1} & \ldots & c_{p-1} & \\ \mu & c_{1} & c_{0} & \ldots & c_{p-2} & \\ \vdots & \vdots & \vdots & \ddots & \vdots & \\ \mu & c_{p-1} & c_{p-2} & \ldots & c_{0} & \\ \end{bmatrix*} \begin{bmatrix*}
K \\ \alpha_{1} \\ \alpha_{2} \\ \vdots \\ \alpha_{p}
\end{bmatrix*}
\end{equation}
where $c_{p}$ is the cross-correlation of the signal $y$ with temporal lag $p$ and $\mu$ is the average of $y$ at time $i$. 

The AR filter diverged and it led to poor accuracy for the estimate $\hat{\lambda}$ in all EuRoC sequences. These results show that the measurements are not temporally correlated and do not follow an autoregressive model. 


Finally, a Kalman Filter has been implemented to estimate $\hat{\lambda}$. The model is
\begin{equation}
\lambda_{k} = a \lambda_{k-1} + w_{k}
\end{equation}
\begin{equation}
z_{k} = h \lambda_{k-1} + v_{k}
\end{equation}
where $a=1$ and $h=1$.

The prediction step is done using
\begin{equation}
\hat{\lambda}_{k|k-1} = a \hat{\lambda}_{k-1|k-1}
\end{equation}
and the a priori variance $p_{k|k-1}$
\begin{equation}
p_{k|k-1} = a^{2} p_{k-1|k-1} + q
\end{equation}
where $q$ is the covariance of the model white noise $w$.

The correction step is calculated as follows
\begin{equation}
k_{k|k} = \frac{h p_{k|k-1}}{h^{2} p_{k|k-1}+r}
\end{equation}
where $k$ is the Kalman gain and $r$ is the covariance of the measurement white noise $v$
\begin{equation}
\hat{\lambda}_{k|k} = \hat{\lambda}_{k|k-1} + k_{k|k}(z_{k} - h \hat{\lambda}_{k|k-1})
\end{equation}
and the a posteriori variance $p_{k|k}$
\begin{equation}
p_{k|k} = p_{k|k-1}(1 - h k_{k|k})
\end{equation}

\section{EXPERIMENTAL RESULTS}

We experimented the proposed method using the sequences from EuRoC dataset \cite{Burri25012016} and the ORB-SLAM algorithm \cite{mur2015orb}. The EuRoC dataset provides eleven sequences recorded by an Asctec Firefly hex-rotor helicopter in two different environments, a room equipped with a Vicon motion capture system and a machine hall. We used the frames from one of the front stereo camera (Aptina MT9V034 global shutter, WVGA monochrome, 20 FPS) to emulate monocular vision and the measurements of the MEMS IMU (ADIS16448, angular rate and acceleration, 200 Hz). The ground truth is measured either by the Vicon motion capture system in the sequences recorded in the Vicon room, or by a Leica MS50 laser tracker and scanner in the machine hall environment. 

In the following, the sequences referred as V1\_01, V1\_02 and V1\_03 were recorded in the Vicon room with configuration of texture 1; the sequences referred as V2\_01, V2\_02 and V2\_03 were recorded in the Vicon room with configuration of texture 2; the sequences referred as MH01, MH02, MH03, MH04 and MH05 were recorded in the machine hall using the Leica system. Note that the trajectory of the UAV is different in each sequence. 

\begin{figure}
   \centering{\includegraphics[width=\columnwidth]{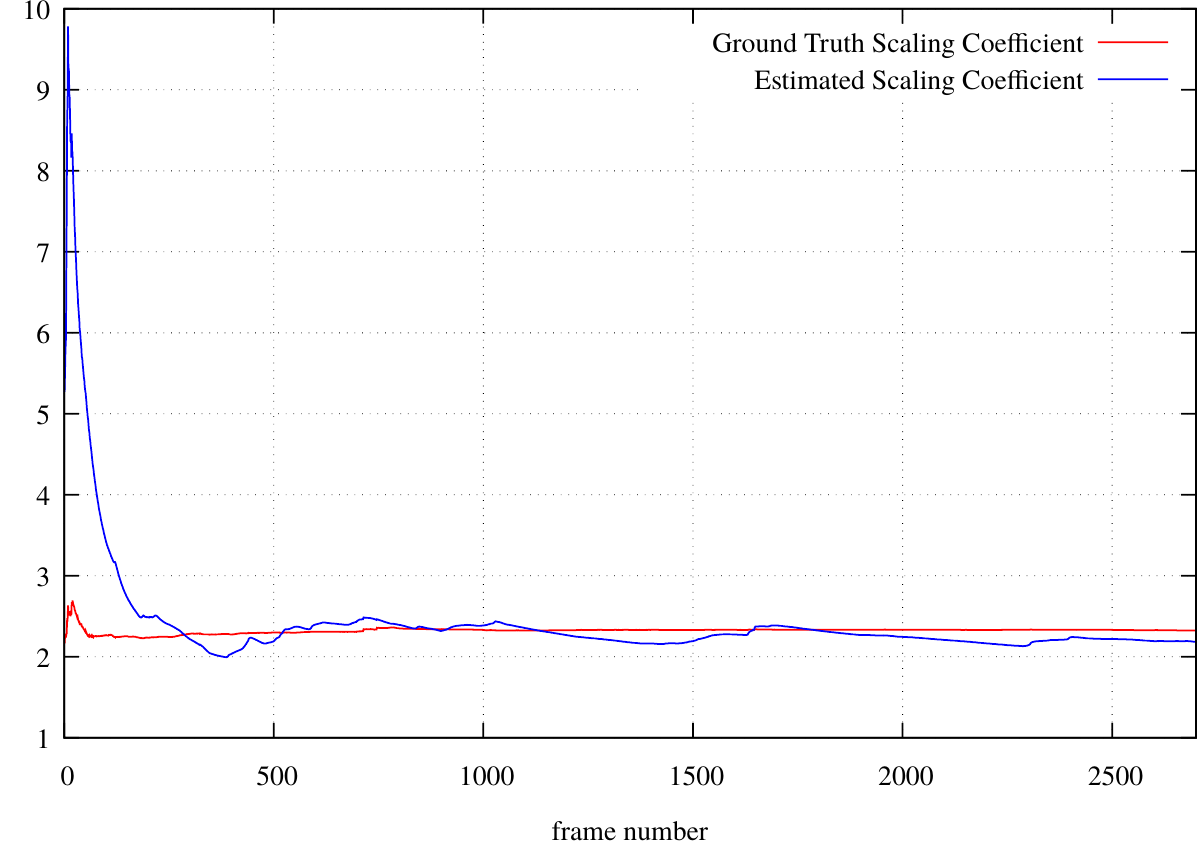}}
   \caption{The estimation of the scaling coefficient $\lambda$ using
     the sequence V1\_01 from EuRoC dataset (In blue the ground truth
     calculated from the Vicon estimates, in red the proposed estimation
     method).}
   \label{scale}
 \end{figure}

The estimation of the scaling coefficient during the sequence V1\_01
is pictured in Figure \ref{scale}. The value of the scaling
coefficient can be compared to a ground truth value, which is computed
using a moving average with Vicon (or Leica) measurements
$^{W\!}\mathbf{t}^{g}$ instead of IMU measurements
 \begin{equation}
 \lambda^{g} = \frac{1}{M} \sum_{k=1}^{M-1} \left( \frac{\Vert ^{W\!}\mathbf{t}^{g}_{F_{k}, F_{k+1}}\Vert_{2}}{\Vert ^{W\!}\mathbf{t}^{\text{m}}_{F_{k}, F_{k+1}}\Vert_{2}} \right)
 \end{equation}
The error $e_{\lambda}$ between the ground truth scaling coefficient $\lambda^{g}$ and the coefficient we estimate using inertial measurement $\hat{\lambda}$ is calculated as follows
\begin{equation}
e_{\lambda} = \Vert \lambda^{g} - \hat{\lambda} \Vert_{1}
\end{equation}
where $\Vert . \Vert_{1}$ is the $L^{1}$ norm.

 \begin{table*}
   \caption{Scaling coefficient and effect on the trajectory}
   \centering{\include{results}}
   \label{table}
 \end{table*}

\begin{figure}
    \centering{\includegraphics[width=\columnwidth]{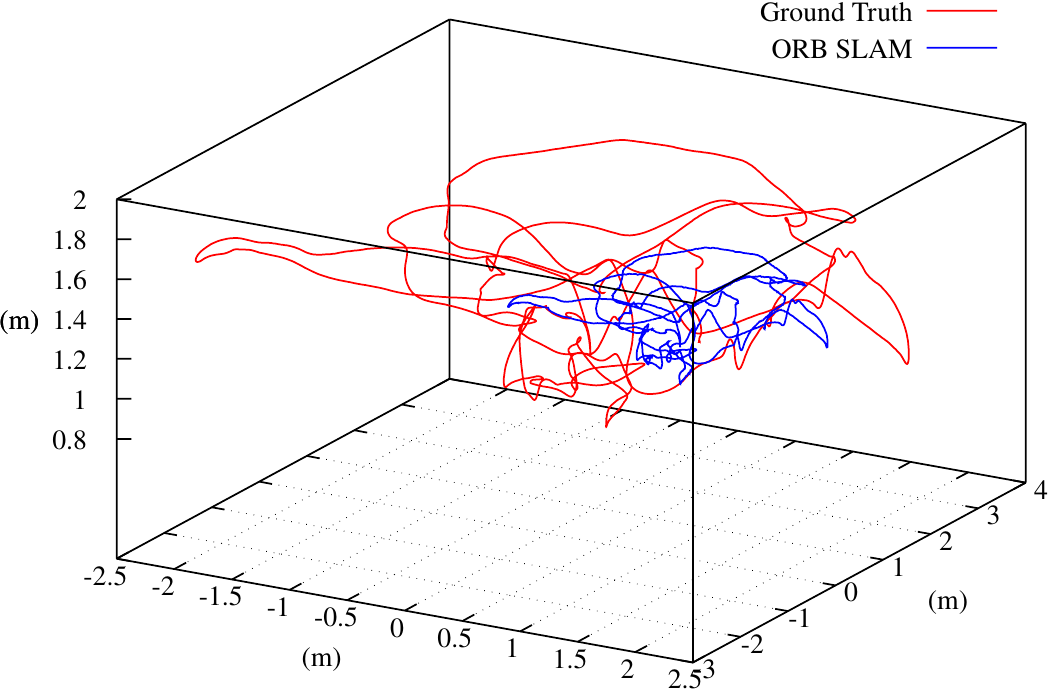}}
    \caption{The trajectory of the camera during the V1\_01
      sequence from EuRoC dataset in the world coordinate frame
      (ORB-SLAM measurements are in blue, the ground truth measured
      with a Vicon motion capture system are in red).}
    \label{traj_not_scaled}
  \end{figure}

\begin{figure}
    \centering{\includegraphics[width=\columnwidth]{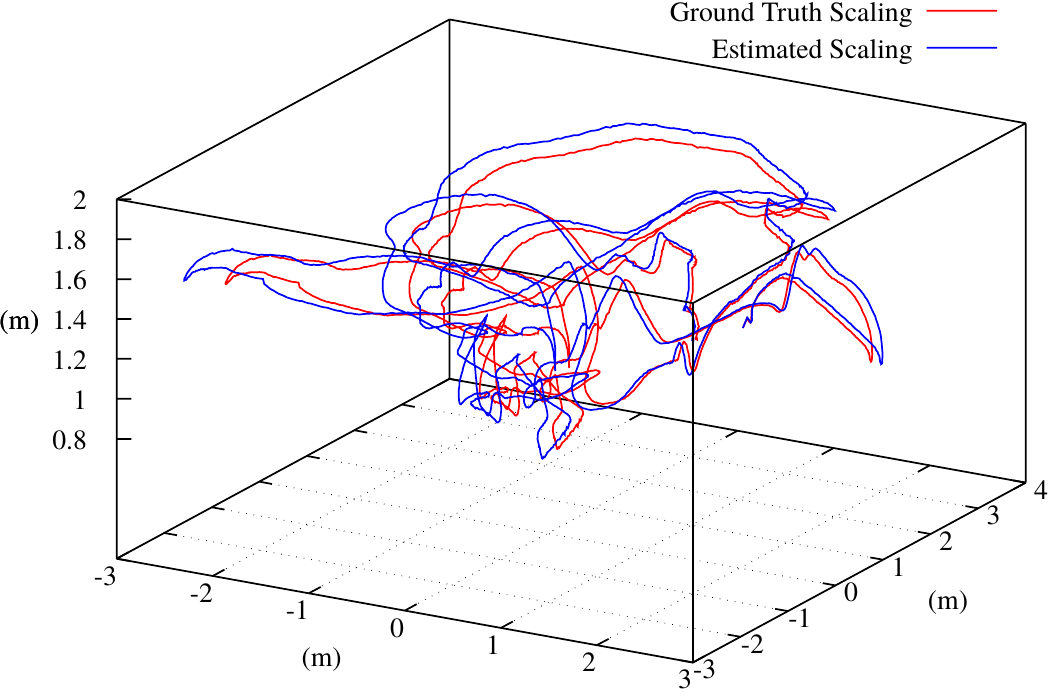}}
    \caption{The rescaled trajectory of the camera provided by
      ORB-SLAM in the V1\_01 sequence from EuRoC dataset in the
      world coordinate frame (the blue trajectory was rescaled using
      the value of $\lambda$, the red trajectory was rescaled using
      the ground truth scaling coefficient).}
    \label{traj_scaled}
  \end{figure}

We rescaled the trajectory provided by the monocular algorithm. As presented in Fig. \ref{traj_not_scaled}, the distances given by the monocular algorithm are arbitrary but consistent, therefore the UAV's trajectory is scaled differently than the ground truth. In Fig. \ref{traj_scaled}, we rescaled the monocular trajectory of the sequence V1\_01 using the ground truth $\lambda^{g}$ and estimated $\hat{\lambda}_{1}$ scaling coefficients. We computed the root-mean-square deviation (RMSE) for each sequence as follows
  \begin{equation}
  \text{RMSE} = \sqrt{\frac{\sum_{i=1}^{M}\left( \lambda^{g}\mathbf{x}_{i}^{T} - \hat{\lambda}\mathbf{x}_{i}^{T} \right) \left( \lambda^{g}\mathbf{x}_{i} - \hat{\lambda}\mathbf{x}_{i} \right)}{M}}
  \end{equation} 
  where $M$ is the number of frames in the sequence and $\mathbf{x}$ is the position of the camera in the world coordinate given by the monocular algorithm (ORB-SLAM for our experiments). The RMSE for each EuRoC sequence is given in Tab. \ref{table}.
  
  The initial trajectory of the UAVs in the sequence V1\_01 provided by the ORB-SLAM algorithm is displayed in Fig. \ref{traj_not_scaled}. The same trajectory rescaled using the scaling coefficients $\lambda^{g}$ and $\hat{\lambda}_{1}$ is displayed in Fig. \ref{traj_scaled}. 
  
\begin{figure}
   \centering{\includegraphics[width=\columnwidth]{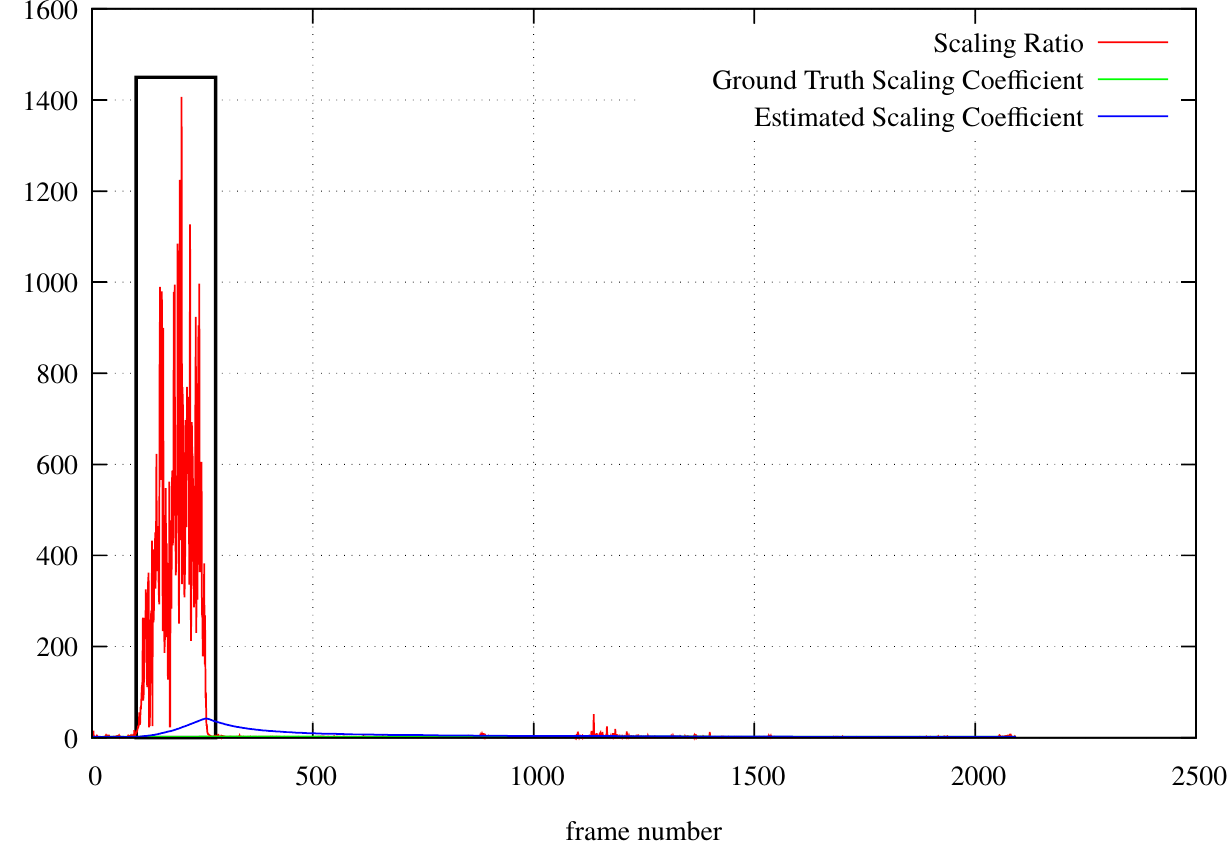}}
   \caption{An example of the corruption of estimates due to the fast translational movements (shown in the rectangular area as marked) for the Leica calibration in one of the sequences recorded in the machine hall environment (MH05).}
   \label{MH_calib}
 \end{figure}

As expected, the bigger the scaling coefficient error $e_{\lambda}$, the bigger the RMSE. The sequences recorded in the Vicon room provide trajectories with lower RMSE than the sequences recorded in the machine hall. The difference between the two sets of sequences can be explained by the strong excitation of the IMU for the calibration of the Leica laser that incorporates a lot of noise in the measurements $\mathbf{t}^{i}$. In the Vicon sequences, the estimate $\hat{\lambda}_{1}$ outperforms. The sequences recorded in the machine hall are very challenging for the inertial fusion because the UAV performed very fast translational movements during a few seconds for Leica ground truth calibration purposes which result in large acceleration measurements and partially corrupt the inertial estimates as shown in Fig. \ref{MH_calib}. Therefore, in the machine hall sequences where strong noise corrupts some IMU measurements, $\hat{\lambda}_{2}$ is far better than $\hat{\lambda}_{1}$, which never managed to completely absorb the strong perturbations of the calibration. Interestingly, Kalman Filter (KF) gives also satisfactory results for Leica sequences. The results of Kalman Filter can be further improved with a finer tuning of the process noise variance $q$. For instance, with a smaller value for $q$, the RMSE of sequences MH\_03 and MH\_04 drops to 0.17 m and 0.76 m respectively. As every single EuRoC sequence is quite different from the others, finding a nice tuning value for the Kalman Filter is not straightforward. We recommend to tune the filter accordingly to the type of flight the UAV performs (smoothness and aggressiveness of trajectories, motion speed, angular velocities). If a Kalman Filter cannot be implemented or tuned, $\hat{\lambda}_{2}$ remains a acceptable estimate. More broadly, keeping the IMU out from large perturbations by using smooth trajectories provides more accurate estimates. The estimates computed through the AR filter, which are not presented because of the dramatically large value of RMSE, show that there is no temporal correlation of the error. 

\addtolength{\textheight}{-13cm}

\section{Concluding Remarks and Future Work}

We presented a fast and easy-to-implement method for the calculation
of the scaling coefficient by fusing inertial measurements with
monocular pose estimation. Monocular camera systems, due to their
nature, can not provide the real-world scale of the pose estimates. To
overcome this problem, we use the inertial measurements produced by an
IMU to i) estimate the scaling coefficient, which relates the
monocular camera pose estimation to the real-world scale, and ii)
speed up the pose estimation by exploiting the availability of the
inertial measurements in very high rates.

The method is highly modular, which makes each component to be easily
replaceable with one's preferences without impacting the overall
operation of the system.


To improve the current method, we plan to further investigate the
initialization of the IMU integration process, particularly with the
incorporation of the current estimate of the scaling coefficient when
appropriate.

We defined three approaches for calculating the scaling coefficient
with regard to the nature of the trajectory followed by the UAV. We
found that the Kalman Filter approach gives accurate estimates when
the tuning is done well, which unfortunately can be hard to do for
some applications. Determination of the tuning value of the process
noise is a complex topic which will be part of our future research
work and experimentations.

\section*{ACKNOWLEDGMENT}

This work was carried out in the framework of the Labex MS2T and
DIVINA challenge team, which were funded by the French Government,
through the program €œInvestments for the Future€ managed
by the National Agency for Research (Reference ANR-11-IDEX-0004-02).

\bibliographystyle{IEEEtran}
\bibliography{bib/filter}

\end{document}

%% file: symbols-notation.tex
\begin{tabular}{ l p{6.5cm} }
  \hline\noalign{\smallskip}
  Notation & ~ ~ ~ ~ ~ ~ ~ ~ ~ ~Description \\
  \hline
  \hline\noalign{\smallskip}
  \{\emph{A}\} & The coordinate frame \{\emph{A}\} referred as $A$ in equations. \\
  $^{A}\mathbf{R}_{B}$ & 3-by-3 rotation matrix that rotates vectors from \{$B$\} to \{$A$\}. \\
  $\mathbf{t}$ & 3-by-1 translation vector. \\
  $\mathbf{t}^{i}$ & Translation vector calculated from inertial measurements. \\
  $\mathbf{t}^{m}$ & Translation vector calculated from monocular vision measurements. \\
  $\mathbf{t}^{g}$ & Translation vector calculated from ground truth measurements. \\
  $^{W}{\textbf{t}}_{F_{i}, F_{j}}$ & Translation vector between the coordinate frames $F_{i}$ and $F_{j}$ written in the world coordinate frame \{\emph{W}\}. \\
  $^{A}\mathbf{T}_{B}$ & Transformation matrix that transforms \{\emph{B}\} into \{\emph{A}\}. \\
  $^{A}\mathbf{T}_{B_{p}}$ & Transformation matrix that transforms \{\emph{B}\} at time $p$ into \{\emph{A}\}, implies that $^{A}\mathbf{T}_{B}$ is changing along time with respect to \{$A$\}. \\
  $F_{i}$ & Camera coordinate frame associated with the image frame $i$. \\
  $\lambda$ & Scaling coefficient. \\
  $\mathbf{b}_{a}$, $\mathbf{b}_{\omega}$ & IMU biases for the accelerometers and gyroscopes. \\
  $\mathbf{g}$ & The gravity vector. \\
  $\mathbf{a}(p)$ & 3-by-1 vector which contains the accelerometer measurements at time $p$ (one vector component per axis). \\
  $\bm\upomega(p)$ & 3-by-1 vector which contains the gyroscope measurements at time $p$ (one vector component per axis). \\
  $\Delta T$ & Time step in seconds. \\
  $^{A}\mathbf{p}$ & 3-by-1 vector of a 3-D measurement in \{$A$\}. \\
  \hline
\end{tabular}%


%% file: results.tex
{\scriptsize
\begin{tabular}{|c||d{3.2}|d{3.2}|d{3.2}|d{3.2}|d{3.2}|d{3.2}|d{3.2}|d{3.2}|d{3.2}|d{3.2}|d{3.2}|d{3.2}|}
\hline
EuRoC      &  \multicolumn{3}{c|}{$\hat{\lambda}$} & \mc{$\lambda^{g}$} & \multicolumn{3}{c|}{$e_{\lambda}$} & \multicolumn{3}{c|}{RMSE (m)} & \multicolumn{2}{c|}{Total distance (m)} \Tstrut \\
\cline{2 - 4} \cline{6 - 13}
Sequence   & \mc{$\hat{\lambda}_{1}$} & \mc{$\hat{\lambda}_{2}$} & \mc{KF} &  & \mc{$\hat{\lambda}_{1}$} & \mc{$\hat{\lambda}_{2}$} & \mc{KF} & \mc{$\hat{\lambda}_{1}$}  & \mc{$\hat{\lambda}_{2}$} & \mc{KF} & \mc{Ground truth} & \mc{Best estimate} \Tstrut \\
\hline
\hline
V1\_01 & 2.49   & 1.89 & 12.42  & 2.31 & 0.19 & 0.42 & 10.11 & \textbf{0}.\textbf{22} & 0.51 & 12.25 & 59.26 & 63.91 \Tstrut\\
\hline
V1\_02 & 1.80   & 1.41 &  7.57 & 2.35 & 0.55 & 0.94 & 5.22 & \textbf{0}.\textbf{55} & 0.95 & 5.25 & 76.58 & 58.74 \Tstrut\\
\hline
V1\_03 & 2.55   & 1.92 &  6.73 & 3.54 & 1.00 & 1.63 & 3.19 & \textbf{0}.\textbf{46} & 0.75 & 1.47 & 77.53 & 55.78 \Tstrut\\
\hline
V2\_01 & 2.92   & 2.33 & 1.17 & 3.02 & 0.10 & 0.70 & 1.86 & \textbf{0}.\textbf{09} & 0.60 & 1.67 & 36.93 & 35.66 \Tstrut\\
\hline
V2\_02 & 4.32   & 1.87 & 82.15 & 3.56 & 0.77 & 1.684 & 78.80 & \textbf{0}.\textbf{47} & 1.03 & 47.99 & 83.92 & 101.93 \Tstrut\\
\hline
V2\_03 & 1.94   & 1.42 & 1.26 & 2.15 & 1.84 & 2.36 & 2.52 & \textbf{1}.\textbf{66} & 2.13 & 2.27 & 139.15 & 71.30 \Tstrut\\
\hline
MH01   & 251.58 & 7.57 & 10.10 & 6.92 & 208.66 & 0.65 & 3.18 & 216.55 & \textbf{0}.\textbf{68} & 3.30 & 85.50 & 93.54 \Tstrut\\
\hline
MH02   & 51.91  & 2.64 & 1.08 & 2.94 & 48.97 & 0.30 & 1.86 & 146.69 & \textbf{0}.\textbf{90} & 5.56 & 84.39 & 75.88 \Tstrut\\
\hline
MH03   & 35.81  & 2.31 & 4.93 & 3.77 & 32.04 & 1.46 & 1.16 & 39.85 & 1.82 & \textbf{1}.\textbf{45} & 131.13 & 171.45 \Tstrut\\
\hline
MH04   & 36.74  & 3.78 & 9.28 & 7.51 & 29.23 & 3.74 & 1.17 & 32.78 & 4.19 & \textbf{1}.\textbf{98} & 103.58 & 127.94 \Tstrut\\
\hline
MH05   & 33.21  & 2.05 & 2.27 & 2.74 & 30.47 & 0.69  & 0.47 & 109.16 & 2.48 & \textbf{1}.\textbf{68} & 116.02 & 96.22 \Tstrut\\
\hline
\end{tabular}
}
